\documentclass[conference]{IEEEtran}
\IEEEoverridecommandlockouts

\usepackage{cite}
\usepackage{amsmath,amssymb,amsfonts}
\usepackage{algorithmic}
\usepackage{graphicx}
\usepackage{textcomp}
\usepackage{eso-pic}
\usepackage{xcolor}

\usepackage[ruled,vlined]{algorithm2e}
\usepackage{array}

\usepackage{flushend}
\newcommand\Mark[1]{\textsuperscript#1}

\def\BibTeX{{\rm B\kern-.05em{\sc i\kern-.025em b}\kern-.08em
    T\kern-.1667em\lower.7ex\hbox{E}\kern-.125emX}}
    
\makeatletter

\usepackage{eso-pic}
\ifCLASSINFOpdf
\else
\fi
\begin{document}

\title{Detection of Bangla Fake News using MNB and SVM Classifier}

\author{\IEEEauthorblockN{Md Gulzar Hussain\Mark{1},
Md Rashidul Hasan\Mark{2}, Mahmuda Rahman\Mark{3}, Joy Protim\Mark{4} and Sakib Al Hasan\Mark{5}}

\IEEEauthorblockA{Department of CSE, Green University of Bangladesh, Bangladesh\Mark{1}\textsuperscript{,}\Mark{2}\textsuperscript{,}\Mark{4}\\
Department of CSE, Jahangirnagar University, Dhaka, Bangladesh\Mark{3}\\Department of CSE, Huzhou Normal University, Huzhou, China\Mark{5}\\
gulzar.ace@gmail.com\Mark{1}, rashidulhasan777@gmail.com\Mark{2}, aurthi018@gmail.com\Mark{3},\\
protimjoy@gmail.com\Mark{4}, sakibgreen29@gmail.com\Mark{5}}}

\maketitle


\begin{abstract}
\boldmath
Fake news has been coming into sight in significant numbers for numerous business and political reasons and has become frequent in the online world. People can get contaminated easily by these fake news for its fabricated words which have enormous effects on the offline community. Thus, interest in research in this area has risen. Significant research has been conducted on the detection of fake news from English texts and other languages but a few in Bangla Language. Our work reflects the experimental analysis on the detection of Bangla fake news from social media as this field still requires much focus. In this research work, we have used two supervised machine learning algorithms, Multinomial Naive Bayes (MNB) and Support Vector Machine (SVM) classifiers to detect Bangla fake news with CountVectorizer and Term Frequency - Inverse Document Frequency Vectorizer as feature extraction. Our proposed framework detects fake news depending on the polarity of the corresponding article. Finally, our analysis shows SVM with the linear kernel with an accuracy of 96.64\% outperform MNB with an accuracy of 93.32\%. 
\end{abstract}



\section{Introduction}
Fake news is an oxymoron that undermines the credibility of reporting that actually meets the verifiability standard and the public interest-i.e. real news. And in this electronic era, it is one of the biggest challenges to control the spreading of false or misleading news due to the free flow of information through social networking sites such as Facebook, Twitter, YouTube, micro-blogging and others. For instance, a new report says, nearly one in three citizens across the United States, Spain, Germany, United Kingdom, Argentina, and South Korea claim they have seen false or misleading information on social media related to COVID-19 \cite{nielsen_fletcher_newman_brennen_howard_2020}.  Because of the abundant spreading of false and unreliable information, some commentators are now referring to the latest wave of misinformation that's accompanied by the COVID-19 pandemic as a disinfodemic \cite{UNESCO}.

Since the information shapes the way we perceive the universe, therefore, fake news and other misleading facts can take on different aspects. Research in intelligence shows that the rumors spreading in social media leave a particularly lasting impact on less smart people \cite{10.1016/j.intell.2017.10.005} and keep them from making the right decisions. False news is used to build people's fears, 
raise racist ideas,
and lead to bullying and violence against innocent people.
Even fake news has great democratic impacts. American presidential election showed how it disrupts and fallacy people's opinions \cite{Grinberg374}.
In the last few years, there have been many tragic incidents in Bangladesh due to rumors. In July 2019, five people were beaten to death and ten injured by mobs as a result of widespread rumors about the expected human sacrifice in the construction of the Padma Bridge \cite{Mobsbeat88:online}. Because of the expanding number of clients in web-based life, news can be immediately distributed by anyone, and its credibility stands traded off. As fake news is written to mislead the readers, it makes it a difficult task to detect based on the content of the news only. It becomes essential to bring an efficient system to detect the fake news as the news content is diverse in terms of style and subject in which it is written. Fake news detection has lately received considerable interest from researchers. So far different approaches have been introduced in the detection of fake news\cite{bondielli_marcelloni_2019}.

Most of the techniques suggested in the literature to identify fake news deal with the problem as a classification task by associating labels such as fake or real, true or false, etc. with a particular text. In most cases, machine learning and deep learning methods are utilized to achieve promising outcomes. Research shows that SVMs have outperformed a number of supervised machine learning algorithms for deception detection in text by exploiting content-based features (e.g. linguistic and visual features)\cite{rubin-etal-2016-fake}\cite{zhang_fan_zheng_liu_2012}. Although a lot of work has been done to identify the fake news, it is still being done in the Bangla Language to a limited extent\cite{sharma2019automatic}\cite{8934841}. With evolving digital technology, the native Bangla speakers are now producing a large amount of Bangla text online. As a result, it will be easier to create a dataset by scraping the text from online sources and categorize them according to polarity of the article by analyzing the text pattern. 

In this work, detection of fake news is proposed using two different modalities available in an efficient manner using Support Vector Machine (SVM) and Multinomial Naive Bayes (MNB). Content-based features have been extracted from the open source data to classify the fake news. Rest of the paper is organized as follows. In section \ref{Rel} we provide an overview of the related works. Section \ref{Medhod} discusses the methodology i.e. how the data are collected and pre-processed, feature extraction method and detection techniques. Section \ref{Res} demonstrates and discusses the result achieved through research and finally Section \ref{con} draws the conclusion with possible future research directions.

\section{Related Works}\label{Rel}
The rising prevalence and vast amount of fake news draws researchers to analyze problems that internet users suffer from. A. S. Sharma et al. proposed a hybrid extraction technique from text documents, combining Word2Vec and TF-IDF\cite{sharma2019automatic} which may detect with standard CNN architecture whether a text document in Bangla is satire or not with a precision of more than 96\%. T. Islam et al \cite{8934841} retrieves comments as data from different social media using comment extractors and removes punctuation marks, numerical meaning, emoticons which create a perfect text corpus. Using Naive Bayes Classifier, which is commonly used for spam identification, they used TF-IDF vectorizer to extract the features from the text corpus and train the processing results.

Some researchers have been utilizing graph analysis to better identify origins of fake news. Shu et al.\cite{Shu2019} have shown that models of network diffusion can be used to map the provenance nodes and the provenance origins of fake news. Tarek Hamdi et al.\cite{10.1007/978-3-030-36987-3_17} suggested an method that incorporates node embedding and user-based functionality to improve Twitter's analysis of fake news. Using node2vec they retrieved information from twitter followers/followees graph which provides a simplified way to help identify SOFNs. The finding that characteristics produced by graph embedding are efficient but node embedding features are powerful SOFN predictors that convey valuable information about a user's reputation in the social network. Kai Shu et al.\cite{shu_mahudeswaran_liu_2018} proposed a model named Social Article Fusion (SAF) which combines the linguistic features of news content with social context features to identify fake news. They use commonly used autoencoders to identify the news content to represent text content in lower dimensional space. To capture the users temporal interactions with the fake news they used Recurrent Neural Networks (RNN), which efficiently performs to capture the temporal relationship. Then they extract useful features after classifying the data, and build various machine learning models to identify fake news.

Mohamed Torky et al.\cite{Torky2019} suggested a novel blockchain-based algorithm named Proof of Credibility (PoC) to identify and prevent fake news and deceptive social media content. The experimental results gave around 89\% accuracy. Eugenio Tacchini et al. has shown that Facebook posts can be categorized as hoaxes or non-hoaxes with high accuracy depending on the users who "like" them. Use user IDs as features for classifying posts, they implemented two ML algorithms (logistic regression and harmonic boolean label crowdsourcing) obtained accuracies exceeding 99\% even with very limited training sets\cite{tacchini2017like}. But, the method offers difficult-to-beat efficiency, its implementation is necessarily limited to situations, because the method uses social interactions (i.e. "likes") as signals to help distinguish Facebook posts, it can not be used when a post has no likes, and it is likely to perform worse when a post only receives few social interactions. 

To detect fake news, Kai Shu et al.\cite{Shu2017ExploitingTF} introduced a novel system called TriFN (which can isolate useful functionality independently from news provider and user obligations, as well as catch interrelationships simultaneously) in his another research work. Marco L. Della Vedova et al.\cite{Vedova2018AutomaticOF} suggested a novel technique of machine learning that incorporates news content and social signals and implemented their plan inside a Facebook Messenger chatbot and accepted it with a legitimate application\cite{tacchini2017like}. For contrast, they achieve higher accuracies than \cite{tacchini2017like} and \cite{Shu2017ExploitingTF} using their respective Facebook and Twitter data sets.

\section{Proposed Methodology}\label{Medhod}
\begin{figure*}[ht]
    \centering
   \includegraphics[width=14cm,height=8cm]{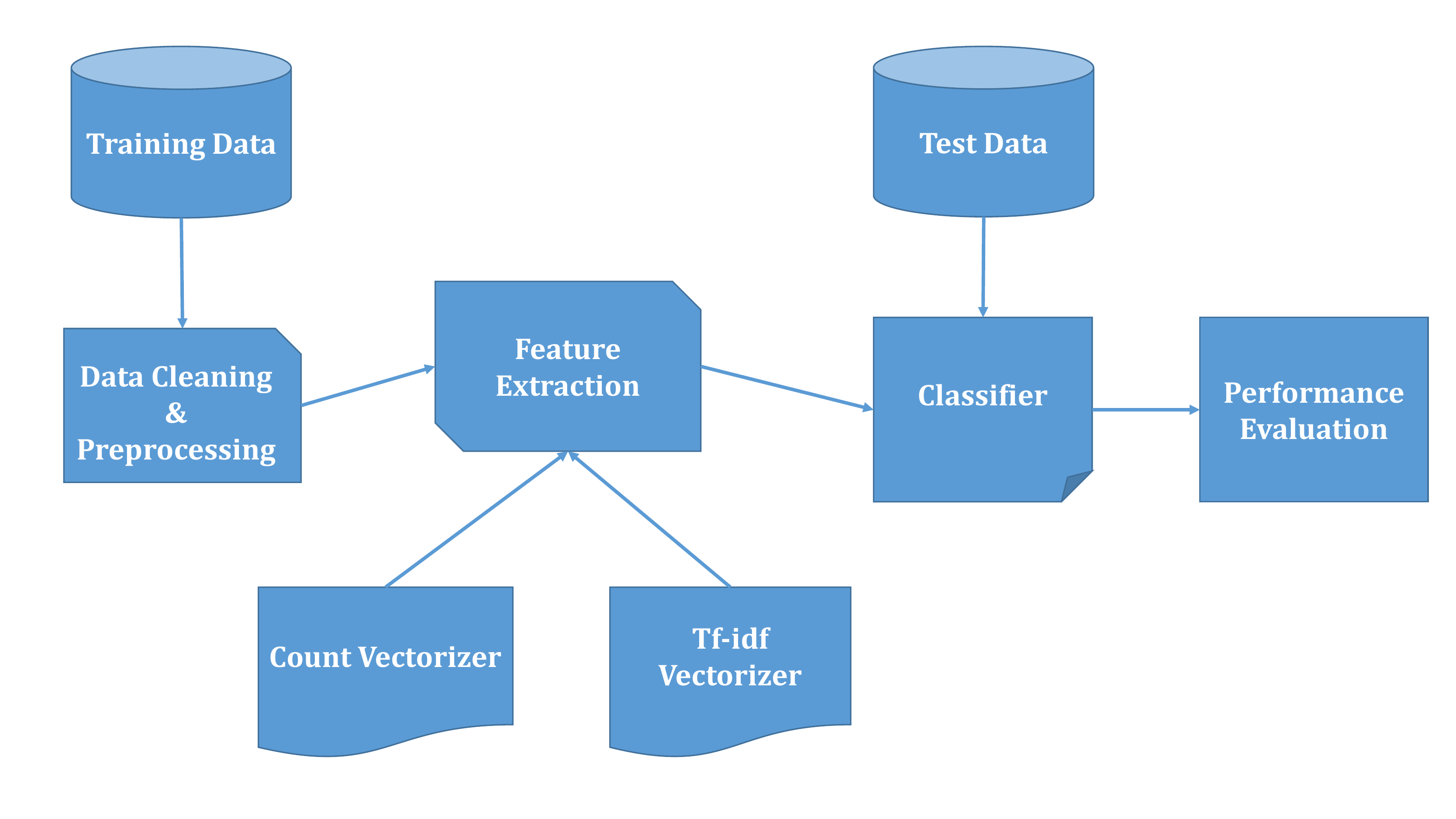}
    \caption{System Flow Diagram of the Proposed System}
    \label{systemFlow}
\end{figure*}

In this work, two classifiers SVM and MNB have been used to classify fake news from Bangla news articles. The dataset used in the proposed classification system has been built by scraping various Bangla newspapers as it is hard to find resources in Bangla language. System flow diagram of the classification model is illustrated in Figure \ref{systemFlow} and others steps are described below.

\subsection{Data Collection}
For the research purpose, we scraped various news articles from ProthomAlo\footnote{https://www.prothomalo.com/}, ProthomAlu\footnote{https://prothom1alu.blogspot.com/},
BalerKontho\footnote{http://www.balerkontho.net/}, Motikontho\footnote{https://motikontho.wordpress.com/} as no previous datasets were found in Bangla language as well as it was hard to gather fake Bangla news. Around 2500 articles were collected for our dataset as all these are public articles. We decided that articles from very popular portals are real news and fake news articles are collected from various sites which are known as portals for satire news such as Motikontho, BalerKnotho, ProthomAlu, etc. Table \ref{table:data} provides the percentage of real and fake news in our dataset.

\begin{table}[htb!]
\centering
\caption{Data set Details Information}

\begin{tabular}{||m{8em} | m{8em} | m{8em} ||} 
 \hline
Type of Article & Total Count & Percentage to Total \\ [0.5ex] 
 \hline
Real & 1548 & 60.92\\
\hline
Fake & 993 & 39.08\\
\hline

\end{tabular}

\label{table:data}
\end{table}

\subsection{Data Preprocessing}
It is very important to apply some preprocessing on the raw text data before feeding them to the classifier. A raw text might contain unnecessary symbols or other things that are not important for our classification. Various emoticons like :D, ;) might be helpful for sentiment analysis but not in our case. We also removed various special characters e.g. @, \#, ! etc from our text. These elements can reduce or diminish the performance of the classifier. After removing the numerical values, punctuation marks, special symbols, our dataset is prepared for the classifier algorithms. Table \ref{table:prepro} shows details characters which were considered removing in preprocessing.

\begin{table}[htb!]
\centering
\caption{Characters Considered removing in Preprocessing}

\begin{tabular}{||m{7em} | m{10em} ||} 
 \hline
  Category & Characters \\ [0.5ex] 
 \hline
 Special Characters & @, \#, \$, \%, |, ,, ......\\
 \hline
 Bangla \& English Digits & 1, 2, 3, 4, .... 0\\
 \hline
 English Alphabets & A, B, C, ...... Z; a, b, c,......z\\
\hline
Emoticons & :), :D, :(, :o, .....\\
\hline

\end{tabular}

\label{table:prepro}
\end{table}

\subsection{Feature Extraction}
Extracting the proper feature have a perfect impact on the performance of the machine learning classifier algorithms. Count Vectorizer and TF-IDF Vectorizer (term frequency–inverse document frequency) are used to extract features from our text before feeding it into the classification algorithms. Count Vectorizer generates a vector which has as several dimensions as the specific word of corpora. Every single word has a particular dimension and contain 1 in that specific dimension with 0 in others which simply keeps the frequency of every words. TF-IDF vectorizer features numerical representations of the words whether they are there or not, rather than just featuring a count. Words are measured by frequency, multiplied by the inverse document frequency of them. In simple terms, words that appear a good amount but everywhere should be provided hardly any significance or weighting. In Bangla Language words like following,

\begin{figure}[ht]
    \centering
   \includegraphics[width=8cm,height=.6cm]{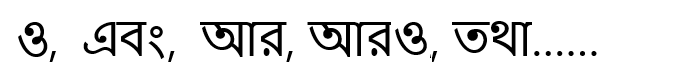}
    \label{OEbong}
\end{figure}

don't offer a huge deal of interest. If a word occurs quite low, or sometimes occurs, then such words are actually more relevant and should be carefully weighed as such. This would lead to improved performance on classification. It is a method intended to describe the significance of a keyword inside a text. If the term denoted as '$t$', a particular document as '$d$' and the whole document as '$D$', then the formula\cite{tf–idfWi99:online} is, 

\begin{equation}
    tf-idf(t, d, D ) = tf (t, d) * idf(t, D) 
\end{equation}
    Here,\\
    \centerline{$tf(t,d)$ is the frequency of '$t$' in '$d$'}\\
    \centerline{$idf(t, D)$ is how '$t$' is common or rare across '$D$'}\\

\subsection{Classifier}
The dataset was split into two sets, training-set and test-set to feed into classifier algorithms. The training set contains 70\% of the dataset and the test set contain 30\% of the dataset. Among various classifier, we used two widely used Multinomial Naïve Bayes Classifier and Support Vector Machine Classifier with a linear kernel in this research. The Naive Bayes classifier is based on Bayes theorem, a simple probabilistic classifier which is fast and easy to implement. It is a tough task concerning which Naïve Bayes version will be implemented. In the context of text classification Multinomial Naïve Bayes gets better results than the Bernoulli Naïve Bayes or Gaussian naïve Bayes\cite{kowsari2019text}. Multinomial Naïve Bayes is often used in a classification problem where the numerous occurrences of the word mean a lot. On the other hand, the support vector machine would be used for regression or classification problems. It utilizes the kernel technique to process the data, and determines an appropriate boundary between the potential outputs depending on these transformations. It is robust against over-fitting problems related to high-dimensional space particularly for text datasets\cite{kowsari2019text}. The linear kernel preferred for text classification as most of them are linearly separable.

\section{Experimental Result}\label{Res}
In this section, performance analysis of multinomial naive bayes and support vector machine classifiers on our dataset has been described. The dataset was split into two parts randomly. One part contains 70\% of the data which is used to train the classifiers and the remaining 30\% is to test the performance of the classifiers.

Multinomial Naive Bayes gave us 93.32\% accuracy for our dataset where Support Vector Classifier with linear kernel gave 96.64\% accuracy. We observed that SVM performs better than MNB. Table \ref{table:resultmnb} and \ref{table:resultsvm} shows the precision, recall and F1-score for MNB and SVM classifier models consecutively.

\begin{table}[htb!]
\centering
\caption{Results for MNB}

\begin{tabular}{||m{5em} | m{5em} | m{5em} | m{5em} ||} 
 \hline
  & precision & recall & f1-score \\ [0.5ex] 
 \hline
Real & 0.90 & 1.00 & 0.95\\
\hline
Fake & 1.00 & 0.83 & 0.90\\
\hline
avg & 0.95 & 0.91 & 0.93\\
\hline

\end{tabular}

\label{table:resultmnb}
\end{table}

\begin{table}[htb!]
\centering
\caption{Results for SVM}

\begin{tabular}{||m{5em} | m{5em} | m{5em} | m{5em} ||} 
 \hline
  & precision & recall & f1-score \\ [0.5ex] 
 \hline
Real & 0.97 & 0.99 & 0.98\\
\hline
Fake & 0.99 & 0.95 & 0.97\\
\hline
avg & 0.98 & 0.97 & 0.97\\
\hline

\end{tabular}

\label{table:resultsvm}
\end{table}

\begin{figure}[ht]
    \centering
   \includegraphics[width=6cm,height=6cm]{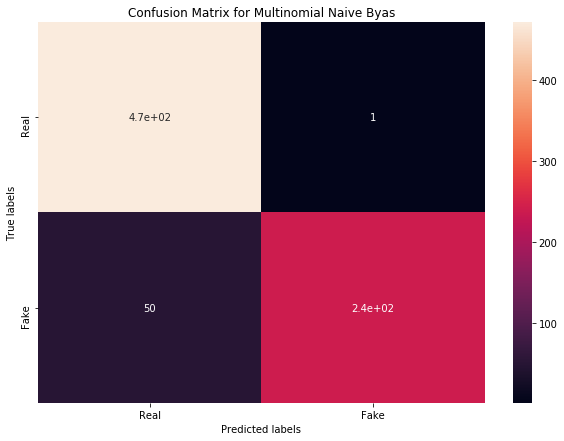}
    \caption{Confusion Matrix for Multinomial Naïve Bayes.}
    \label{MNBCM}
\end{figure}

Figure \ref{MNBCM} shows the confusion matrix for Multinomial Naïve Bayes which shows that 472 news was indicated as real and was actually real news. On the other hand 50 news was predicted as fake which was actually real, that means false negative. Similarly, 240 fake news were predicted as fake and was actually fake but only 1 was predicted as fake and was actually real which new was falsely indicated as real.  

\begin{figure}[ht]
    \centering
   \includegraphics[width=6cm,height=6cm]{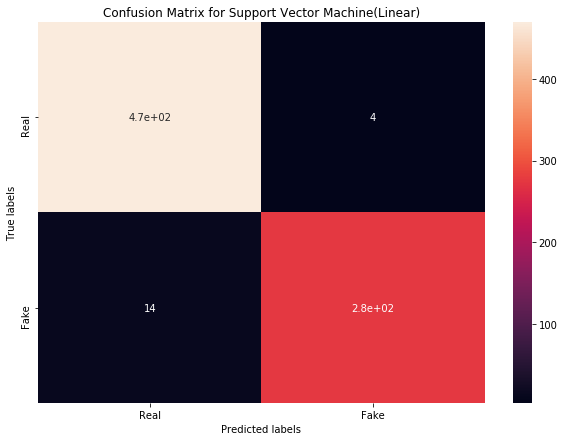}
    \caption{Confusion Matrix for Support Vector Machine with Linear Kernel.}
    \label{SVCCM}
\end{figure}

Figure \ref{SVCCM} shows the confusion matrix for Support Vector Classifier which shows that 469 news was indicated as real and was actually real news. On the other hand only 14 news was predicted as fake which was actually real, that means false negative. So we can see SVC gives less false negative. Similarly, 276 fake news were predicted as fake and was actually fake but only 4 was predicted as fake and was actually real which new was falsely indicated as real.

\section{Conclusion}\label{con}
This research work concludes that Support Vector Machine with linear kernel is doing marginally better than Multinomial Naïve Bayes on our dataset. Since the research with the identification of Bangla fake News is new\cite{sharma2019automatic}\cite{8934841}, this work will be helpful for any future works on such topic. Very recently a research work has been accepted in a conference consisting a dataset around 49 thousand news articles\cite{hossain2020banfakenews} and it is our plan for the future to work with this dataset to expand the number of features and sufficient lexicons. Besides, a stemmer can be applied to minimize corpus size and enhance model efficiency. In the future, the rate of success can be improved by doing more research utilizing hybrid-classifiers.

 

\ifCLASSOPTIONcaptionsoff
  \newpage
\fi

\bibliographystyle{IEEEtran}

\bibliography{ref.bib}

\end{document}